\documentclass[runningheads,a4paper]{llncs}
\usepackage[T1]{fontenc}

\usepackage{amsmath}
\usepackage{amsfonts}
\usepackage{amssymb}
\usepackage{graphicx}
\usepackage{array}

\usepackage{hyperref}
\usepackage{color}

\usepackage{chngcntr}
\counterwithout{figure}{section}
\counterwithout{figure}{subsection}

\newcolumntype{R}[1]{>{\raggedleft\let\newline\\\arraybackslash\hspace{0pt}}m{#1}}

\urldef{\mailsa}\path|{jan.wietrzykowski, michal.nowicki, piotr.skrzypczynski}@put.poznan.pl|
\newcommand{\keywords}[1]{\par\addvspace\baselineskip
\noindent\keywordname\enspace\ignorespaces#1}

  \usepackage{everypage} 
\newcommand\atxy[3]{%
 \AddThispageHook{\smash{\hspace*{\dimexpr+#1\relax}%
  \raisebox{\dimexpr+\voffset-#2\relax}{#3}}}}

\begin{document}

\atxy{1cm}{-1cm}{\parbox{\textwidth}{\footnotesize{The final publication is available at Springer via \url{http://dx.doi.org/10.1007/978-3-319-54042-9_58}}}}

\mainmatter  

\title{Adopting the FAB-MAP algorithm for indoor localization with WiFi fingerprints}

\titlerunning{Adopting the FAB-MAP algorithm for indoor localization \ldots}

%
%
\author{Jan Wietrzykowski\thanks{Corresponding author} \and Micha\l{} Nowicki \and Piotr Skrzypczy\'{n}ski}
\authorrunning{Jan Wietrzykowski \and Micha\l{} Nowicki \and Piotr Skrzypczy\'{n}ski}

\institute{Institute of Control and Information Engineering,\\
           Pozna\'n University of Technology\\
           ul. Piotrowo 3A, 60-965 Pozna\'n, Poland\\
\mailsa\\}

%
%

\toctitle{Lecture Notes in Computer Science}
\tocauthor{Authors' Instructions}
\maketitle

\begin{abstract}
Personal indoor localization is usually accomplished by fusing information from
various sensors. A common choice is to use the WiFi adapter that provides information
about Access Points that can be found in the vicinity. Unfortunately, state-of-the-art
approaches to WiFi-based localization often employ very dense maps of the WiFi signal
distribution and require a time-consuming process of parameter selection. On the other hand,
camera images are commonly used for visual place recognition, detecting whenever the user
observes a scene similar to the one already recorded in a database. Visual place
recognition algorithms can work with sparse databases of recorded scenes and are
in general simple to parametrize. Therefore, we propose a WiFi-based global localization
method employing the structure of the well-known FAB-MAP visual place recognition algorithm.
Similarly to FAB-MAP, our method uses Chow-Liu trees to estimate a joint probability
distribution of re-observation of a place given a set of features extracted at places visited so far.
However, we are the first who apply this idea to recorded WiFi scans instead of visual words.
The new method is evaluated on the UJIIndoorLoc dataset used in the EvAAL competition,
allowing a fair comparison with other solutions.

\end{abstract}

\section{Introduction}
\vspace{-1mm}
Solving the self-localization problem is important for mobile robots, as the
key component of autonomy. On the other hand, localization is also crucial for people,
allowing them to find themselves in an unknown environment, and then efficiently
navigate to the destination. Outdoor localization is mainly a solved problem since
the introduction of the Global Positioning System (GPS). Unfortunately, the GPS signal
is not available indoors, and therefore various solutions are developed to allow
similar functionality in buildings. Especially interesting are solutions that do
not require modifications to the existing infrastructure of buildings, allowing to
introduce personal localization into existing sites (e.g. office buildings, shopping malls)
easily and at a reasonable cost. Among the variety of existing approaches,
the solutions relying only on the sensors of the ubiquitous mobile devices
(e.g. smartphones) are the most appealing to the potential users. Nowadays,
WiFi signals are available in most buildings that might be of interest for
personal localization, whereas those signals can be received by every modern
mobile phone or tablet. Thus, WiFi-based localization is of high practical importance.
WiFi localization based on signal fingerprints is similar to visual place
recognition when it comes to the underlying principles of data processing.
Both classes of localization systems contain a pre-recorded database (called radio map for WiFi),
and the current perception (image or signal scan for WiFi). Those data are
compared to entries in the database to find similarity between the current
perception and the places observed in the past.
Therefore, we propose to adopt the statistical learning mechanism by means of the
Chow-Liu tree \cite{chowliu}, which was proven successful in the FAB-MAP algorithm to WiFi
fingerprints used as features that describe unique places. We believe that the
resulting method should be able to better understand the influence of information
contained in the appearance of WiFi Access Point (AP) in the scan for localization
purposes, than the already known algorithms for WiFi-based indoor localization.

\section{Related work}
\vspace{-1mm}
Typical mobile devices are equipped with a variety of sensors:
accelerometers, gyroscopes, magnetometers, cameras and WiFi/Bluetooth adapters.
Due to the limited precision of each sensing modality, the best localization
results are obtained when information from different sensors is fused.
The inertial sensors are usually combined to create Attitude and Heading
Reference System (AHRS)~\cite{ieeesensors2014}, which outputs the orientation of the device.
The camera is also a useful sensor, as it enables to perceive the surroundings
in a way similar to human perception. Processing a continuous stream of images
allows the localization system to estimate pose by means of Visual Odometry (VO),
which can be computed in real-time on a mobile device~\cite{iciar14},
but the process is computationally demanding and thus has a negative impact on the battery life.
Alternatively, the images can be processed at discrete poses to detect if the
camera observes an already visited location or a place from a pre-recorded
database describing the environment~\cite{ipin16nasze}.
Visual place recognition systems allow obtaining a global position estimate
without much computational burden. The FAB-MAP algorithm~\cite{fabmap} is a
state-of-the-art solution proven in challenging environments. It learns the
frequency of occurrence of similar (visual) features and co-occurrence of
those features in order to obtain robust place recognition results.

In WiFi-based localization, the state-of-the-art is WiFi fingerprinting~\cite{radar},
which assumes that it is possible to obtain a precise and dense map of scans in
known locations (radio map). During the localization phase, a WiFi scan is compared
to the scans stored in the radio map, finding best matches according to the chosen metrics.
To obtain the final position of the user, the positions of $k$ best matches
(nearest neighbors) from the radio map are averaged without or with additional weights,
thus this algorithm is called kNN or wkNN~\cite{radar}.
Moreover, when the localization task assumes the existence of multiple buildings and/or floors,
the WiFi-based method estimates at first the building, next the floor, and then performs
the localization procedure described above. The fingerprinting approach with careful,
manual tuning achieves precise localization results, as can be seen in the annual
localization challenge at the Indoor Positioning and Indoor Navigation (IPIN) conference~\cite{ipin15comp}.
The lengthy and sometimes cumbersome process of manual tuning in the fingerprinting
approach encourages researchers to look for a machine learning solution to the
WiFi-based localization. The existing machine learning approaches to WiFi fingerprinting
include using Gaussian processes~\cite{iros16gaussian}, neural networks~\cite{wifineuralnet},
and random forests~\cite{ipin16randomforest}. So far these approaches provide
worse localization results than the best wkNN-based solutions. However, we have
already demonstrated~\cite{2014jamris} that combining WiFi fingerprinting and visual
place recognition can be beneficial, as those systems can be joined to operate in
locations, where one of the systems might fail due to the lack of WiFi signals
or due to the limited visual information being available. This paper extends this
line of research investigating if a machine learning approach successfully used
in visual place recognition may be directly adopted to WiFi fingerprints.

\section{Adopting FAB-MAP to WiFi features}
\vspace{-1mm}
FAB-MAP~\cite{fabmap} was designed especially to perform loop-closing for SLAM (Simultaneous
Localization and Mapping) in robotics, and therefore was developed
with scalability and robustness in mind. The algorithm is based on a probabilistic
model that describes the probability $p(L_i | \mathcal{Z}^k)$ that an already visited
place $L_i$ is being observed, given a set of all feature vectors $\mathcal{Z}^k=\{Z_1,Z_2\ldots,Z_k\}$
extracted at places visited so far (including the present one). This model enables to
determine whether a new place is being visited, or a place visited in the past is re-observed.
Although such interpretation of the model is suitable for the above-mentioned loop-closing
problem, the task of estimating user position for indoor localization requires a different view.
If we substitute the collection of already visited places with a database of known
locations pin-pointed to a floor map, we get a probability distribution of the user
position in this map. This idea is exploited in our recent work~\cite{ipin16nasze}.

\begin{figure}[thpb!]
\centering
 \includegraphics[width=0.8\columnwidth]{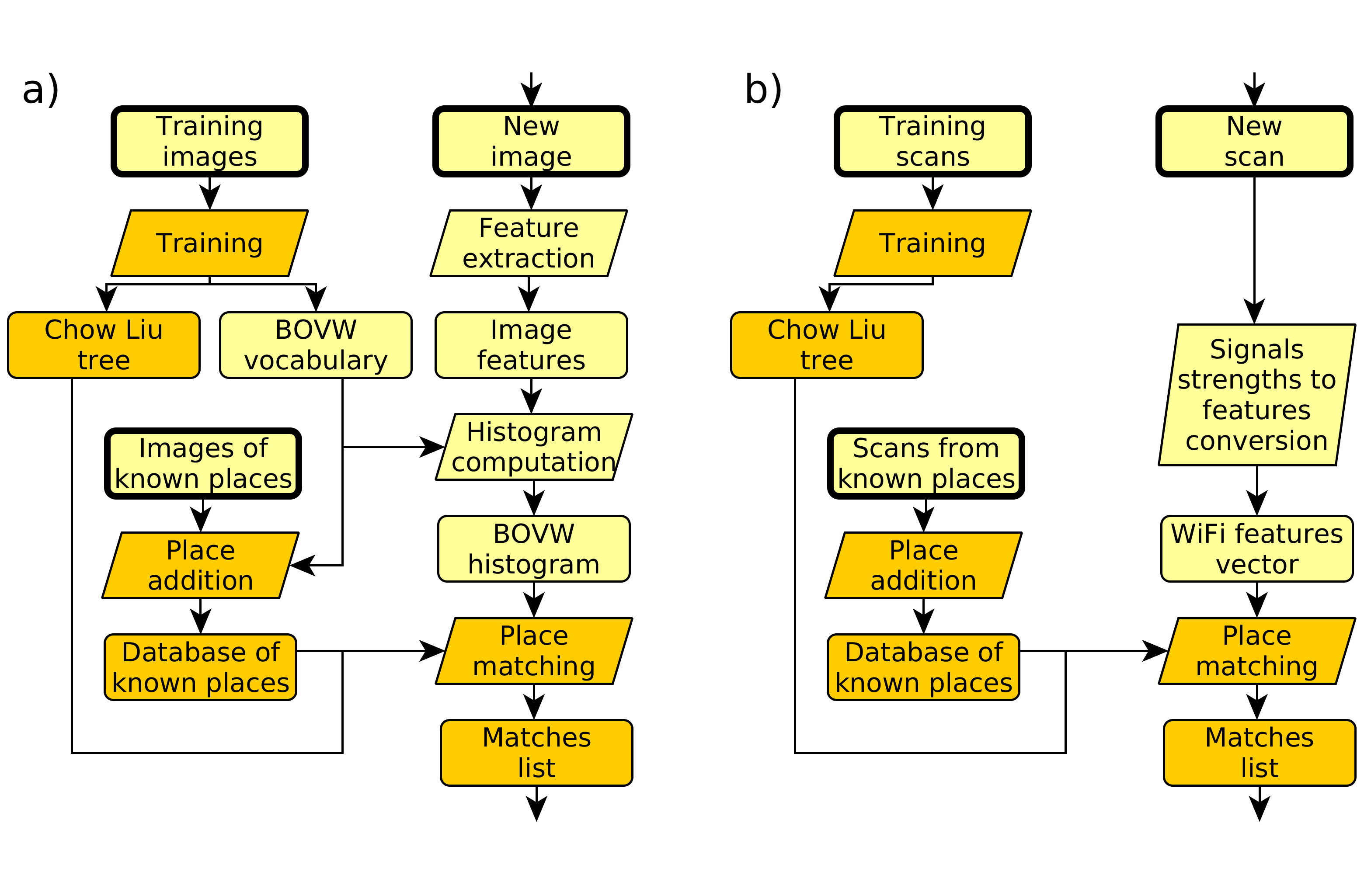}
 \vskip-4mm
 \caption{A schematic overview of a) the original FAB-MAP algorithm,
          and b) the WiFi-based version. Blocks with thick borders
          are input data sources. Blocks denoted by darker color are
          common for both solutions. Note that our version creates feature vectors directly,
          thus it does not use any vocabulary. Therefore, no counterpart to the ``BOVW vocabulary''
          block is present}
 \label{fig:fabmapScheme}
\end{figure}

Our solution is based on the FAB-MAP and exploits the probabilistic inference
mechanism implemented in the original algorithm. The differences come from the
entirely different characteristics of locations we exploit: WiFi signal scans
instead of point-like visual features. This caused substantial changes to the
approach, which is used to create the vector of features, and to the training
phase of the algorithm. A comparison of the block schemes for both algorithms is
depicted in Fig. \ref{fig:fabmapScheme}.
In the FAB-MAP algorithm, the Bag of Visual Words (BOVW) representation of visual (point)
features is used to characterize $j$-th image/location $Z_j$. The BOVW vector is a histogram
of visual words occurrences in an image, where the vocabulary is learned by clustering
salient features extracted from training examples.
The probabilistic model associates each histogram bin with a random variable $z_i \in Z_j$.
If the $i$-th bin is zero, which means that the feature wasn't detected on the image,
then $z_i$ is set to 0, in the opposite case $z_i$ is set to 1.
To separate a location model from the model of dependencies between features the
hidden variables $e_i$ were introduced. These variables indicate whether a feature
is present in the image. Note that the presence of a feature does not imply detection of this feature,
because detectors are not perfect \cite{jamris}. It is not uncommon that features
are not detected despite their presence and that they are detected while being absent.

\begin{figure}[thpb!]
\centering
 \includegraphics[width=0.8\columnwidth]{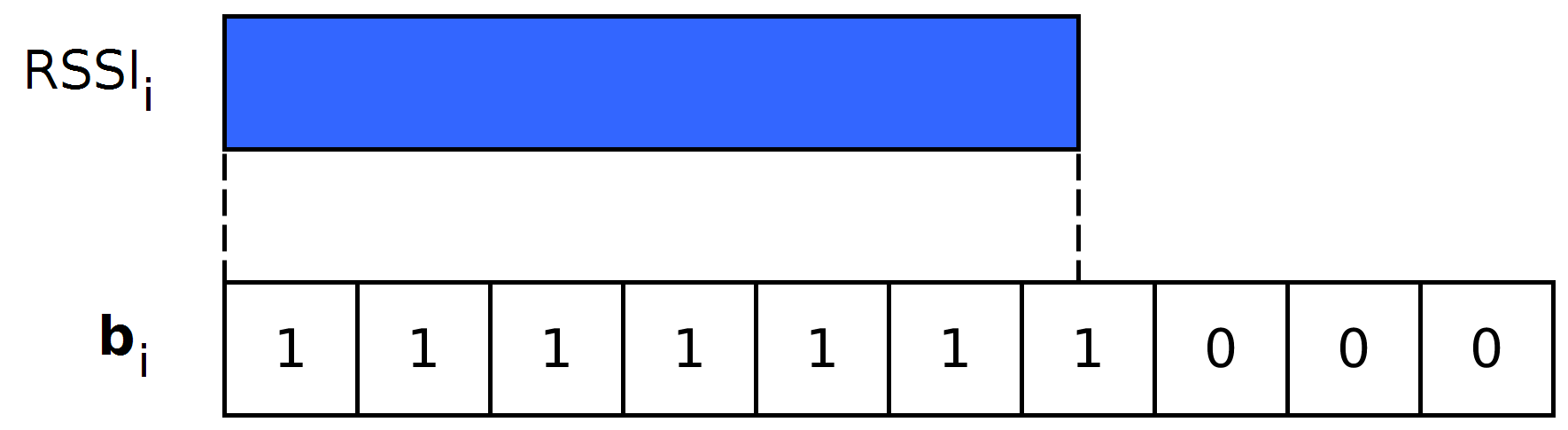}
 \caption{Illustration of conversion from RSSI values to $\mathbf{b}$ vector. RSSI values are expressed in dBm}
\label{fig:featExt}
\end{figure}

The main idea of place recognition based on WiFi scans is to employ vector
features related to the WiFi signal measurements (scans) in place of the
BOVW vectors (histograms). Since FAB-MAP maps each feature onto a binary representation,
a proper conversion of a WiFi scan was necessary to get the most of the descriptive
power out of it. A natural idea that comes to mind is to construct a vector $\mathbf{v}$
of the networks' presence in the current scan with the value $v_i = 0$ indicating
an absence of the $i$-th network, and $v_i = 1$ indicating presence of this network.
This vector should contain $v_i$ for every network, identified by its unique
BSSID number, present in the considered area. Unfortunately, such a conversion
throws away all information about the strength of the signal (RSSI), which may be
beneficial to the ability to discriminate different locations with a similar
pattern of networks presence. Thus, we decided to extend the vector $\mathbf{v}$,
so for every possible network, it contains $k$ bins forming a sub-vector $\mathbf{b}_i$
for $i$-th network. Each bin has an associated threshold and whenever the signal strength
(expressed in dBm) of the $i$-th network exceeds this threshold, its value is set to 1.
Thresholds were uniformly distributed in range $<-110, -10>$ with a step (bin width) of 10 dBm.
If a network is not present in the current scan, then all bins contain 0. This
process is illustrated in Fig. \ref{fig:featExt}.

\begin{figure}[thpb!]
\centering
 \includegraphics[width=0.8\columnwidth]{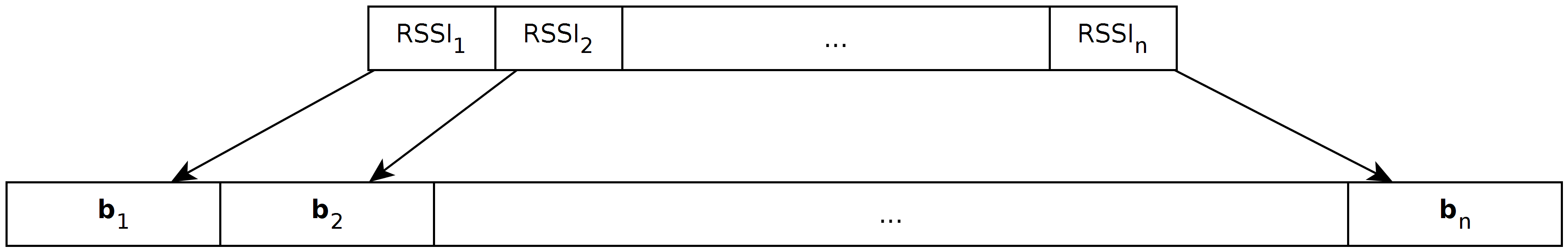}
 \caption{A schematic view of the feature vector extension process}
\label{fig:featExtWhole}
\end{figure}

The final feature vector is a concatenation of all $\mathbf{b}_i$ vectors.
A schematic overview of the feature vector extension is presented in Fig. \ref{fig:featExtWhole}.

The FAB-MAP algorithm requires training phase prior to operation.
The training phase consists of the Chow-Liu tree training step and construction
of a database of known places that are going to be recognized.
Given a dataset of scans made at various places in which multiple scans are 
taken at the same location, we had to split it into two datasets: one
characterizing the environment and containing scans taken at disjoint locations
for Chow-Liu tree training, and another one containing scans that will form a
database of known places.
We used DBScan~\cite{dbscan} algorithm with parameters $\epsilon = 1$ and
$\text{MinPts} = 1$ to divide the input dataset into clusters of scans taken at one location.
The $\epsilon$ parameter bounds the distance in which neighbors are searched,
while $\text{MinPts}$ denotes the minimum number of scans that are allowed to form a cluster.

A random scan drawn from each cluster was used to characterize the environment,
whereas 10 other scans (also drawn at random) were used to construct the database of
known places. It was necessary to use only 10 scans captured at a single location in
order to balance the dataset, because at some locations there were many scans,
while at others only a few. It caused locations with more scans to be favoured
over locations with fewer scans.

A proper place recognition can take place after the training phase.
Given a list of detected networks with BSSIDs and signal strengths for each of them,
a feature vector is computed and feed to the FAB-MAP-like recognition procedure.
The procedure returns a list of matches with a probability assigned to every 
location from the database of known places. We assume that the location with the
highest probability is the current user location in the map.

OpenFABMAP~\cite{openfabmap}, a publicly available C++ implementation of the
FAB-MAP algorithm was used as the basis for our place recognition system.
We made our code also publicly available through the GitHub platform
\footnote{\url{https://github.com/LRMPUT/WiFi-FAB-MAP.git}}.

\section{Experimental data}
\vspace{-1mm}
\label{sec:data}
Another important part of designing a machine learning system is to properly
evaluate the proposed solution. Of no less importance is the ability to compare
the results to the state-of-the-art approaches. Therefore, we decided to use the
UJIIndoorLoc~\cite{uji} dataset, which is a database of WiFi scans captured at
three buildings of Universitat Jaume I in Madrid, Spain. The area at which
scans were taken covers almost 11000 m$^2$ and contains 13 different floors.
A wide range of devices (25) and multiple users guarantee a diversity of scans origin.
The database consists of 19937 training and 1111 validation examples.
For each example, information about detected WiFi networks along with their
signal strength in dBm, longitude, latitude, building ID and floor number is attached.
This information enabled us to set up an experiment to evaluate the efficiency of
our system. Unfortunately, the original testing examples used in the EvAAL competition
at IPIN 2015~\cite{ipin15comp} are not publicly accessible, therefore we used a
part of the training dataset in the validation procedure, and treated the validation
dataset as a test set. It is worth noting that the training dataset was collected
approximately 4 months prior to the collection of validation and testing datasets. Moreover, the datasets were
collected by different users and using different mobile devices. All that makes
the estimation of user location a challenging task, especially taking into consideration
that we were forced to use similar datasets in both the training and validation steps,
which often brings a risk of overfitting the parameters.

\section{Parameter tuning}
\vspace{-1mm}
\label{sec:param}
An important property of machine learning algorithms is an ability to automatically
estimate parameter values used in the algorithm on the basis of training examples.
Nevertheless, usually there are few meta-parameters that have to be set prior to
the training procedure. 

\begin{figure}[thpb!]
\vskip-8mm
\centering
 \includegraphics[width=0.8\columnwidth]{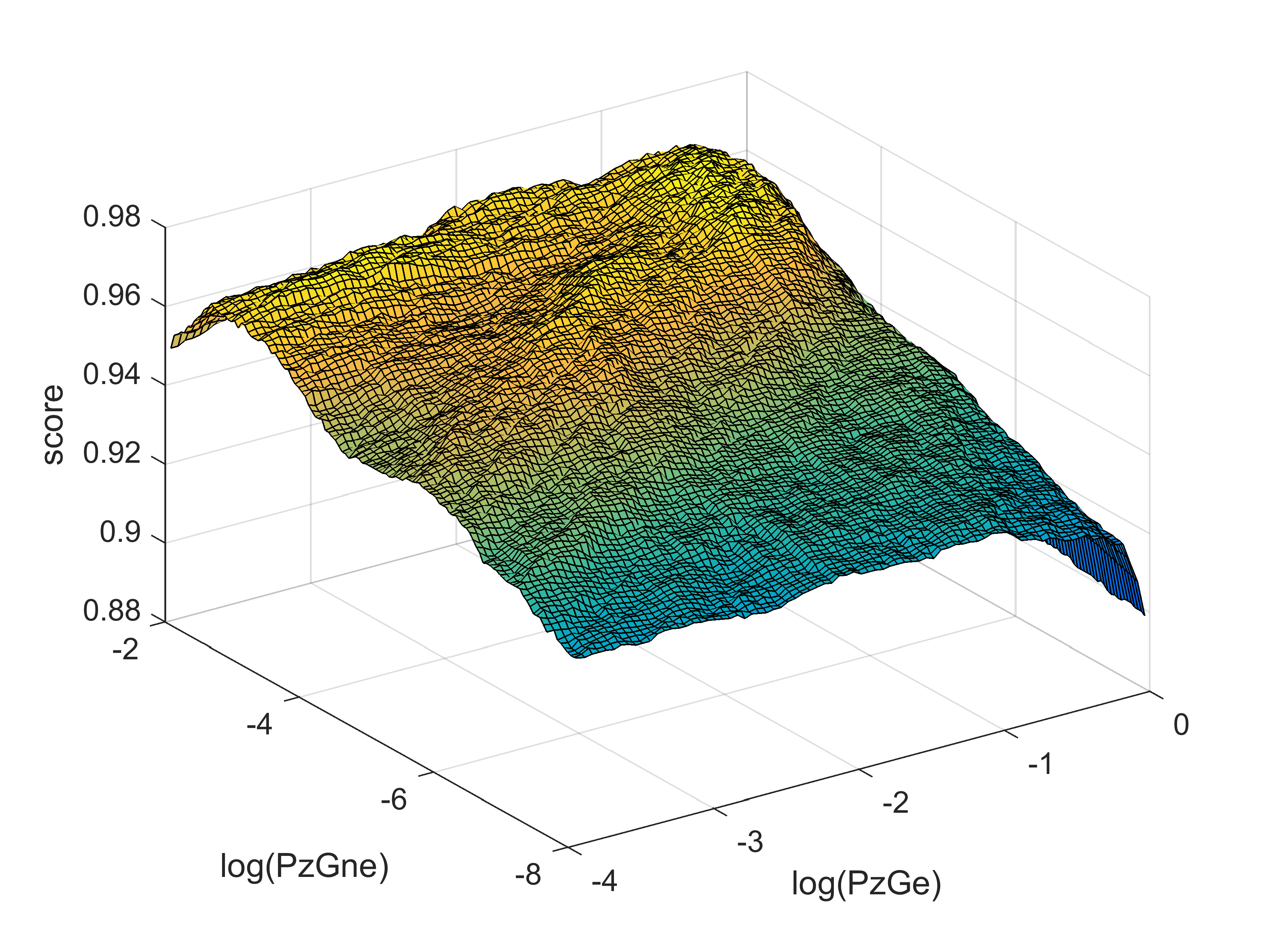}
 \vskip-4mm
 \caption{Score on a separated part of training dataset as a function of parameters.
          Note that parameter axes are logarithmic}
\label{fig:val}
\end{figure}

Although these meta-parameters are often set manually by
the system designer using his experience and observing the system behaviour, we
have decided to automatize this process by using grid search and the validation dataset.
In the FAB-MAP algorithm, there are two parameters that have a substantial influence
on the system behaviour, namely PzGe and PzGne. These are detector model parameters
and determine the $p(z_i = 1 | e_i = 0)$, and $p(z_i = 0 | e_i = 1)$ quantities.
The balanced training dataset containing 10 scans for each location was divided
into two parts: subtraining and validation. The subtraining set was used to construct a
database of known places while the validation one was used to evaluate the current parameter values.
The space of parameters was extensively explored by using grid search with an exponential step.
The PzGe parameter values were swept from $\exp(-0.01) \approx 0.99$ to $\exp(-
4) \approx 0.018$ with a step of exponential function argument equal to $0.05$
and PzGne values from $\exp(-2) \approx 0.14$ to $\exp(-8) \approx 0.00034$ with
the same step size. The use of exponential steps is motivated by a wide-spread
application of the exponential family in probabilistic modeling \cite{expFam}.

In the evaluation procedure, we used the accuracy of predicting a correct floor
and a correct building at the same time as the score measure. Such prediction of
the user location may seem imprecise, but it is often the case when coarse positioning
is done by one algorithm and the fine position is computed by another one.
The results of the parameter tuning procedure are presented in Fig. \ref{fig:val}.
Since there are only two parameters, it was convenient to plot a surface to
visualize the influence of these parameters on the recognition accuracy.
Finally, the best parameters were determined as PzGe = 0.31 and PzGne = 0.043.

\section{Experimental evaluation}
\vspace{-1mm}
\label{sec:exp}
During experimental evaluation we computed score for every tested configuration.
The score was the accuracy of correctly predicting the building and the floor at the
same time. The prediction was considered correct if a building ID and floor number
of the matched location from the database of known places were the same as the
building ID and floor number of the example. Additionally, we calculated the
mean distance error $e_d$ among the correctly classified examples using formula:
\begin{equation}
 e_d = \frac{1}{|\mathcal{C}|} \sum_{i \in \mathcal{C}} |p_i - m(i)|,
\end{equation}
where $\mathcal{C}$ is a set of correctly classified examples, $p_i$ is a 
position of $i$-th example in meters, and $m(i)$ is a position of the
matched location from database of known places.

\begin{table}
 \centering
  \caption{Accuracy results for predicting the building and the floor}
  \label{tab:res}       
  \begin{tabular}{r R{1.5cm} R{1.5cm} R{1.5cm} R{1.5cm}}
    \hline\noalign{\smallskip}
    Bin width & PzGe & PzGne & score & $e_d$ [m] \\
    \hline\noalign{\smallskip}
    5 & 0.3135 & 0.0429 & 0.82 & 9.99 \\
    5 & 0.3135 & 0.0043 & 0.89 & 8.50 \\
    5 & 0.3135 & 0.0004 & 0.92 & 8.55 \\
    10 & 0.4916 & 0.0550 & 0.81 & 9.96 \\
    10 & 0.4916 & 0.0055 & 0.89 & 8.21 \\
    10 & 0.4916 & 0.0006 & 0.91 & 8.40 \\
    \noalign{\smallskip}\hline
  \end{tabular}
\end{table}

\begin{figure}[thpb!]
\vskip-8mm
\centering
 \includegraphics[width=0.9\columnwidth]{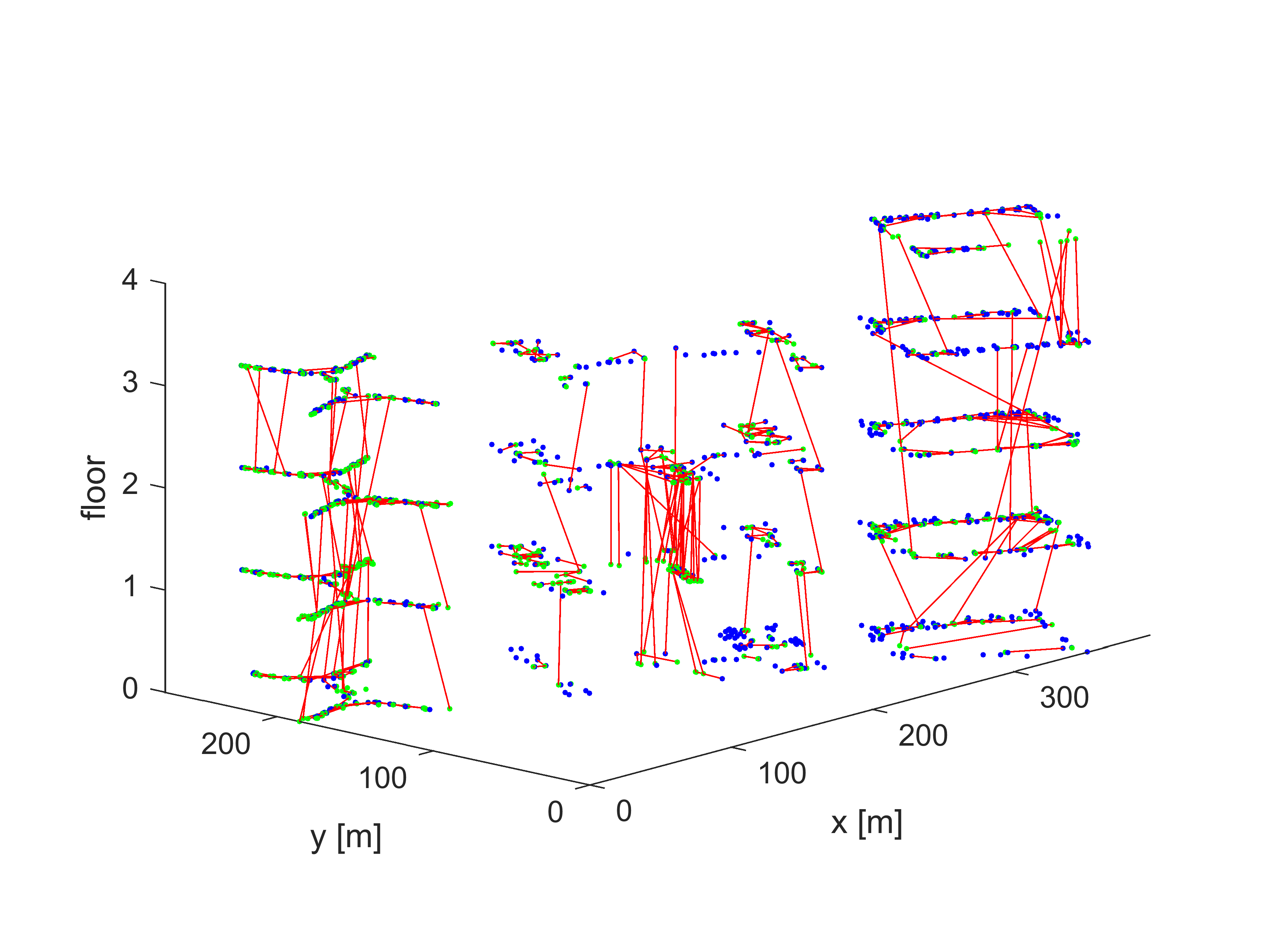}
 \vskip-6mm
 \caption{Visualization of location assignments. Blue points are locations from database of known places,
          green points are testing examples, and red lines are matches between testing examples and locations from the database of known places}
\label{fig:assig}
\end{figure}

The results of the experimental evaluation are gathered in Tab. \ref{tab:res}.
We tested parameters obtained in a validation procedure as well as parameters
with lowered PzGne values. The decision to test lower values of PzGne was motivated
by the fact that in the case of WiFi scans it is rather unlikely to not detect a network
that is in the range of a device, and we were not in a possession of a proper dataset
for the validation procedure.
In Fig. \ref{fig:assig} we plotted location matches for all examples from the testing dataset.
Red lines connect locations of examples with matched locations from the database of
known places. Most of the matches are invisible because they are very short, but the
rest indicate that frequent mistakes are floor mismatches. Locations that are above
or below the true location are prone to be wrongly matched, which is expected as
the WiFi signal easily penetrates the ceiling separating floors.

\section{Conclusions}
\vspace{-1mm}
\label{sec:con}
We proposed and tested a novel WiFi fingerprinting method which adopts the FAB-MAP algorithm,
originating from the visual appearance-based place recognition.
The key part of the algorithm adoption is the new method of feature vectors generation from a
list of detected WiFi networks. Another relevant contribution of this paper is the procedure for
preparation of the training data for the Chow-Liu tree and for the database of known places.
The solution was evaluated on a challenging, publicly available dataset and proved to provide
satisfactory results. To fully exploit the automated pipeline of the system tuning a proper
validation dataset would be needed. Only by examining the score on examples that were collected
independently, under different conditions the parameter tuning procedure would be enabled to find
parameter values that are not overfitted.
As it comes to the influence of bin width, the difference between dividing the signal strength
range into 22 (bin width of 5) or 11 (bin width of 10) intervals can be neglected, justifying
the use of larger bins for computation efficiency.

Future work will focus on locating the user or robot using WiFi data clusters much smaller than whole floors.
Knowledge of the corridor in which a device is located often can be sufficient for indoor navigation
when combined with other premises about the user position \cite{mobicase}.
We also plan to implement a more detailed location model than the one available in OpenFABMAP.

\section*{Acknowledgements}
\vspace{-2mm}
This research was funded by the National Science Centre in Poland
in years 2016-2019 under the grant 2015/17/N/ST6/01228. This work
was also partially supported by the Pozna\'n University of Technology grant DSPB/0148

\bibliographystyle{unsrt}
\bibliography{biblio}

\end{document}